\documentclass[10pt,twocolumn,letterpaper]{article}

\usepackage{iccv}
\usepackage{times}
\usepackage{epsfig}
\usepackage{graphicx}
\usepackage{amsmath}
\usepackage{amssymb}
\usepackage{booktabs}
\makeatletter
\@namedef{ver@everyshi.sty}{}
\makeatother
\usepackage{tikz}
\usepackage{multirow}

\newcommand{\eoe}[0]{\mathcal{E}}

\newcommand{\mysubsubsection}[1]{\par{\emph{\bf #1}}}

\newenvironment{enumerate*}%
  {\begin{enumerate}%
    \setlength{\itemsep}{0pt}%
    \setlength{\parskip}{0pt}}%
  {\end{enumerate}}
  
 \newenvironment{itemize*}%
  {\begin{itemize}%
    \setlength{\itemsep}{0pt}%
    \setlength{\parskip}{0pt}}%
  {\end{itemize}}
  
\usepackage[breaklinks=true,bookmarks=false]{hyperref}

\iccvfinalcopy 

\def\iccvPaperID{****} 

\ificcvfinal\fi
\begin{document}

\title{JEDI: Joint Expert Distillation in a Semi-Supervised Multi-Dataset Student-Teacher Scenario for Video Action Recognition}

\author{Lucian Bicsi$^1$, Bogdan Alexe$^{1,*}$, Radu Tudor Ionescu$^1$, Marius Leordeanu$^{2,3}$\\
$^1$University of Bucharest, $^2$Politehnica University of Bucharest\\
$^3$Institute of Mathematics of the Romanian Academy\\
$^*${\tt\small bogdan.alexe@fmi.unibuc.ro}
}

\maketitle
\ificcvfinal\thispagestyle{empty}\fi

\begin{abstract}
   We propose JEDI, a multi-dataset semi-supervised learning method, which efficiently combines knowledge from multiple experts, learned on different datasets, to train and improve the performance of individual, per dataset, student models. Our approach achieves this by addressing two important problems in current machine learning research: generalization across datasets and limitations of supervised training due to scarcity of labeled data. We start with an arbitrary number of experts, pretrained on their own specific dataset, which form the initial set of student models. The teachers are immediately derived by concatenating the feature representations from the penultimate layers of the students. We then train all models in a student-teacher semi-supervised learning scenario until convergence. In our efficient approach, student-teacher training is carried out jointly and end-to-end, showing that both students and teachers improve their generalization capacity during training. We validate our approach on four video action recognition datasets. By simultaneously considering all datasets within a unified semi-supervised setting, we demonstrate significant improvements over the initial experts.
\end{abstract}


\section{Introduction}
Modern algorithms tackling specific computer vision tasks in image and video understanding rely on models trained on large scale visual datasets. While the size of representative datasets for different visual tasks has increased over the past years, ranging from tens of thousands in semantic segmentation (MS COCO~\cite{lin2014microsoft}) and action recognition (Kinetics~\cite{K400}) to millions in object recognition (ImageNet~\cite{deng2009imagenet}), they are only able to capture a small part of the complexity of the real world.
In the supervised case, each dataset consists of a set of class labels and a limited set of samples drawn from the infinitely large space of examples in the real world. Usually, the representation obtained based on this data sampling has a strong built-in bias \cite{Torralba11unbiasedlook}, with various factors such as manual example selection, image acquisition, label distribution biases contributing to the overall bias.
In object recognition, a simple linear SVM classifier performs much better than random chance in distinguishing between images of different known datasets \cite{Torralba11unbiasedlook} with similar object classes. In video action recognition, in both ActivityNet \cite{ActivityNet} and UCF101 \cite{UCF101} datasets, ``playing piano'' is the only class depicting pianos \cite{li2018resound}, so a piano detector is sufficient to correctly classify the respective action. Similarly \cite{li2018resound}, classifying scenes as basketball court or soccer field is enough to correctly classify action classes ``basketball dunk'' and ``soccer juggling''. 
Each dataset provides a unique view of the visual world through the data samples it contains. Thus, methods have a hard time achieving generalization across datasets. Indeed, an algorithm trained on a dataset might not perform well on others, as data follows different distributions. Moreover, class labels vary across datasets, and consequently, with each new dataset, a method has to solve a new task.
Creating datasets that properly capture generic patterns, without inheriting biases, constitutes one of the biggest challenges nowadays \cite{Li_2019_CVPR, TommasiPCT17}. 

In this paper, we come to address these challenges, with a model that simultaneously learns from multiple datasets in a semi-supervised fashion.
More precisely, we leverage information from multiple datasets to improve performance on each dataset in part. The main idea is to facilitate understanding the visual world depicted by each particular dataset by using the out-of-distribution data from other datasets. We demonstrate the usefulness of our method for the case of video action recognition. We leverage the use of expert models pretrained on different datasets and group them together in an ensemble of experts. Each expert is specialized in recognizing specific action classes and comes with its own perspective, as learned from the corresponding dataset.  We show that by combining the knowledge of all experts into ensembles of experts (teachers) and employing semi-supervised knowledge distillation (by using the output of teachers as pseudo-labels) back into the individual experts (students), we improve the performance of each expert. Our framework unifies multi-dataset and semi-supervised learning into a single pipeline, being trained jointly and end-to-end. This approach is novel in the literature and proves its effectiveness by improving the student networks over the initial counterparts, at no extra cost during test time.



We conduct experiments on four action recognition datasets: ActivityNet \cite{ActivityNet}, HMDB51 \cite{HMDB51}, Kinetics400 \cite{K400}
and UCF101 \cite{UCF101}.  
Our results show that
we significantly improve the performance of the students (experts) themselves (between $1\%$ and $8\%$).


In summary, we make the following contributions:
\begin{itemize}
\item \vspace{-0.18cm} We propose a novel semi-supervised multi-dataset model for action recognition in video, which learns to combine multiple experts (one per dataset) to create semi-supervised teachers for the next generation of students, over multiple iterations. We address the limited labeled data problem through self-supervision: students at one iteration become teachers for the students at the next iteration. 
\item \vspace{-0.18cm} We address the issue of dataset bias by  simultaneously learning on multiple datasets. We perform tests using four challenging datasets and show that our learning pipeline significantly improves the performance of the individual experts. To our best knowledge, this represents a novel training pipeline.
\item \vspace{-0.18cm} We make learning efficient such that both the distillation of teacher knowledge into the student and learning of the teacher ensembles are carried out jointly and end-to-end. To our best knowledge, this is also novel.
\end{itemize}

\section{Related Work}

\mysubsubsection{Relation to video action recognition methods.} State-of-the-art action  recognition  models use  different  deep  network  architectures  based  on  optical flow~\cite{Feichtenhofer2016CVPR, Simonyan2014nips}, 3D convolutions~\cite{Carreira_2017_CVPR,Hara_2018_CVPR}, or recurrent connections~\cite{Singh_2016_CVPR}. 
These architectures consider as input either a frame or a clip (a set of frames) that are densely \cite{Simonyan2014nips} or randomly \cite{TSN} sampled. They alternatively employ some smart frame or clip sampling strategy \cite{ Gowda2021AAAI,TSM}, learn to select the best frame \cite{Ren_2020_WACV} or use the entire video \cite{Liu_2021_CVPR} by using a cumulative temporal clustering algorithm based on the Hamming distance. In contrast, our method is designed to integrate several expert models pretrained on different datasets, and thus, it can use any of the mentioned models. In particular, we use the Temporal Shift Module \cite{TSM} and the Temporal Segment Network \cite{TSN} architectures in our experiments.

\mysubsubsection{Relation to ensemble and teacher-student learning.} 
Ensembles are widely used in machine learning \cite{Hansen1990pami}. The usual approach in building ensembles is to combine the prediction of different models trained on the same dataset. Distinct from the common methodology, we build an ensemble of expert models which are pretrained on different datasets. Moreover, instead of just using the output of the expert models, we use the representation in the form of hidden features provided by the individual models. Combining experts towards guiding the learning through ensembles has previously been demonstrated for image retrieval \cite{Douze11cvpr}, video retrieval \cite{Liu2019a} or multi-task scene understanding \cite{Leordeanu21aaai}. 
Different from previous work, our approach creates ensembles by forming two-hop pathways that pass through an intermediate representation obtained from all experts trained on their specific dataset and then return to the target expert. In this manner, our multi-task system is in fact a set of ensembles in which all knowledge from all datasets is jointly used.
We combine experts to build ensembles following a teacher-student learning paradigm~\cite{ba14nips,hinton15nipsw}. Usually this is done in a two-step iterative scheme by alternating the ensemble learning of a teacher from students with the learning of students to mimic the teacher. Different from the conventional approach, we formulate the learning of both students and teachers in a joint manner, where both teachers and students improve over each training iteration. 

\mysubsubsection{Relation to multi-task learning.} Multi-task learning was
successfully applied in scene understanding \cite{Leordeanu21aaai, lu2020corr}, video anomaly detection \cite{Georgescu-CVPR-2021}, universal/generic 3D representation \cite{zamir2016generic} or image classification and depth prediction~\cite{doersch2017multitask}. The works of~\cite{Leordeanu21aaai, lu2020corr} exploit consistency between various different tasks such as semantic segmentation, depth and motion, while \cite{doersch2017multitask,Georgescu-CVPR-2021,zamir2016generic} employ a shared feature extraction backbone and train multiple heads for each separate task. In our case, we impose consistency between different datasets, depicting different action classes by retraining the specific experts on the output of ensembles that effectively combine features from all datasets. 

\mysubsubsection{Relation to unsupervised and semi-supervised representation learning.}
Many recent methods that include an unsupervised learning component are based on some form of clustering \cite{caron2020unsupervised, cluster-herbert,unsup-clust-cvpr}, using pretext tasks \cite{predict-rotation, shuffle-herbert, colorization,split-brain} or training 
adversarial generative models \cite{bigbigan}. In our case, the unsupervised learning part is based on training on pseudo-ground-truth labels, which we obtain from the output of our multi-dataset ensembles. This is different from the recent neural graph consensus model \cite{Leordeanu21aaai}, which trains on pseudo-labels from ensembles, but there is no multi-dataset and transfer learning aspect. Different from \cite{Leordeanu21aaai}, we train students and teachers jointly end-to-end.

\section{Proposed Method}

Consider a set of $n$ models $\{\mathcal{M}_1, \mathcal{M}_2,...,\mathcal{M}_n\}$, where each model $\mathcal{M}_i$ is pretrained on samples (and labels) of some dataset $\mathcal{D}_i$. The classes from any two datasets may overlap or be completely disjoint. One may reason that each model in the set $\left\{\mathcal{M}_i\right\}_{i=1}^n$ provides a different perspective for the same data sample, which is induced by the inherent bias of dataset $\mathcal{D}_i$. For simplicity, we will refer to the initial experts $\mathcal{M}_i$ simply as \textit{``experts''}. We aim to improve each expert $\mathcal{M}_i$ by employing self-supervised learning from out-of-distribution knowledge from all other experts $\mathcal{M}_j$, $\forall j \neq i$.

Each expert $\mathcal{M}_i$ is composed of a pretrained encoder model (\eg convolutional layers), and a classification head (\eg fully connected layers). We create $n$ ensembles $\mathcal{E}_i$ (mixtures) of experts, such that there is one ensemble $\mathcal{E}_i$ per dataset. The input to each of the $n$ ensembles is a combined  representation of the sample. We obtain the joint representation by aggregating (via concatenation) the intermediate representations given by all the experts for the given sample. The intermediate representations are taken just before the classification heads. Each ensemble model will ultimately yield outputs (\ie prediction logits) that are similar to the classification head of the corresponding expert model.

\subsection{Alternating vs. Joint Training}

\noindent \textbf{Alternating training:}
Perhaps the most straightforward procedure for training is to start a two-step student-teacher iterative learning process: 1) The ensembles (teachers) learn to classify the data samples more robustly by combining the knowledge of the  students; 2) The experts (students) learn from the teachers via knowledge distillation, by using the output of the teachers as pseudo-labels during training.

\noindent \textbf{Joint training (proposed):} The aforementioned two-step training process can be further improved in terms of learning speed and efficiency: instead of iteratively retraining the students and teachers from scratch, we propose to jointly train both teachers and students in an end-to-end differentiable pipeline, such that with each gradient step in the learning process (per batch), both students and teachers are jointly optimized. 

\subsection{JEDI Training Procedure}

As motivated above, we employ a training procedure to increase the performance of both students and teachers, formulated as a \textit{joint} multi-task semi-supervised teacher-student learning scenario. We specifically identify $3n$ tasks that need to be jointly optimized:

\begin{enumerate*}
    \item Supervised training of each $\mathcal{M}_i$ to classify samples from dataset $\mathcal{D}_i$, for all $i = \overline{1, n}$ ($n$ tasks);
    \item Supervised training of each teacher $\mathcal{E}_i$ to classify samples from dataset $\mathcal{D}_i$, for all $i = \overline{1, n}$ ($n$ tasks);
    \item Unsupervised knowledge distillation of teachers $\mathcal{E}_i$ into their corresponding students $\mathcal{M}_i$ on samples from \textbf{all} datasets $\left\{\mathcal{D}_j\right\}_{j=1}^n$, by using the output predictions of $\mathcal{E}_i$ as pseudo-labels, for all $i = \overline{1, n}$ ($n$ tasks).
\end{enumerate*}

\begin{figure}[!th]
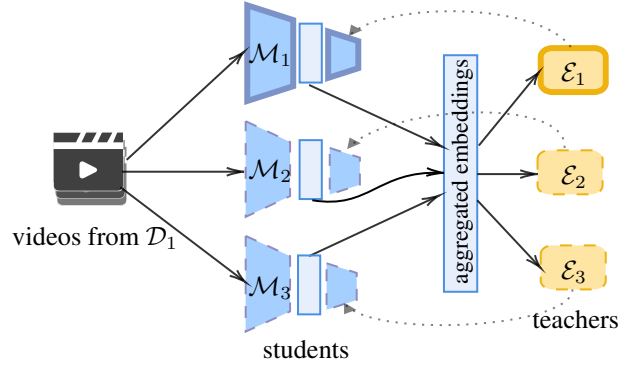

    \centering
    \include{eoe-diagram}
    \caption{For simplicity, we exemplify the iterative JEDI training procedure for $\mathcal{D}_1$. Dashed arrows indicate knowledge distillation.}
    \label{fig:eoe_diagram}
    \vspace{-0.3cm}
\end{figure}

An illustration of the training procedure is shown in Figure \ref{fig:eoe_diagram}, for $n = 3$ models. Each student $\mathcal{M}_i$ is composed of an encoder part and a classification head. We extract intermediate representations just after the final encoder layer, and aggregate them via concatenation. Dashed arrows show how knowledge is distilled in our framework. All models benefit from unsupervised knowledge distillation.


\noindent \textbf{Combined loss criterion.}
The whole end-to-end pipeline is trained in a multi-task scenario, by employing a combined loss for all $3n$ tasks. As knowledge distillation involves all datasets, and classification involves just one dataset $\mathcal{D}_i$ for model $\mathcal{M}_i$, it follows that each model gets to ``see'' more data distributions in the distillation scenario than in the classification one, which helps each teacher and student to generalize better. However, since the number of samples in each dataset $\mathcal{D}_i$ can vary from one dataset to another, the interplay between different data distributions might have negative effects from smaller datasets. We address this misalignment by weighting the distillation loss by a factor of $w_i$. Our novel combined loss for an example $x \in \left\{\mathcal{D}_i\right\}_{i=1}^n$ and its corresponding ground-truth label $y$ is given by:
\begin{equation}\label{eq_loss}
\begin{split}
 \mathcal{L}(x, y)  & = \alpha \cdot \mathcal{L}_{\textit{cls}}(\mathcal{M}_{i}(x), y) + \beta \cdot \mathcal{L}_{\textit{cls}}(\mathcal{E}_{i}(x), y) + \\ 
 & + \gamma \cdot \sum_{j = 1}^{n} w_{ij} \cdot   \mathcal{L}_{\textit{kd}}(\mathcal{M}_j(x), \mathcal{E}_j(x)),
\end{split}
\end{equation}
where $\alpha$, $\beta$ and $\gamma$ are hyperparameters controlling the importance of various loss components, and $w_{ij}$ is a dataset weighting factor defined as follows:
\begin{equation}\label{eq_weight}
w_{ij} = \left\{\begin{array}{ll} 1, & \mbox{if} \; i = j \\ \frac{|\mathcal{D}_j|}{\sum_{k=1}^n |\mathcal{D}_k|}, & \mbox{if} \; i \neq j \end{array}\right. , \forall i=\overline{1,n},\; j=\overline {1,n}.
\end{equation}
Note that we decide to set $w_{ij} = 1$ in Eq.~\eqref{eq_weight} for samples coming from the dataset of the corresponding model, \ie when $i=j$, as we argue that samples from the reference dataset are more important and should impose a greater weight.

In Eq.~\eqref{eq_loss}, $\mathcal{L}_{\textit{cls}}$ can be any desired classification loss function (\eg Hinge loss, cross-entropy loss, etc.), and $\mathcal{L}_{\textit{kd}}$ is a chosen knowledge distillation loss criterion (\eg soft-label MSE, cross-entropy loss, etc.). Moreover, $\alpha$, $\beta$ and $\gamma$ are hyperparameters, and should be chosen empirically according to the task (\eg by cross-validation). The total loss of the joint model is defined as the sum of the combined losses over all training samples of all datasets. Note that for the samples where ground-truth labels are not available (the unsupervised learning phase), only the third term of the loss is active, which essentially performs the student-teacher distillation. However, when ground-truth labels are available (the supervised learning phase), it is important to use them as well (by activating the first two terms), in order to avoid catastrophic forgetting \cite{kemker2018measuring} of the ground-truth labels, thus preserving the stability of the multi-task process and improving the convergence rate. 


\subsection{Weight Freezing and Embedding Caching}

In order to make the training process more efficient, during our experiments, we freeze the encoder architecture of all student models, and only keep the classification heads trainable. This is, undoubtedly, a compromise. If we are to allow the whole architecture to be trained, this may lead to even higher accuracy gains. However, introducing this compromise improves the experimental training time by considerable margins, allowing us to pre-compute the intermediate representations by running the inference pipeline once for all models and all datasets. The intermediate embeddings are persisted on disk. This makes the training procedure essentially equivalent to end-to-end fine-tuning with frozen encoders. 

\begin{figure}[t]
    \begin{center}

\tikzset{every picture/.style={line width=0.75pt}} 

\begin{tikzpicture}[x=0.75pt,y=0.75pt,yscale=-1,xscale=1]

 \draw  [color={rgb, 255:red, 74; green, 144; blue, 226 }  ,draw opacity=1 ][fill={rgb, 255:red, 74; green, 144; blue, 226 }  ,fill opacity=0.14 ] (18,19) -- (25.23,19) -- (25.23,76) -- (18,76) -- cycle ;
\draw  [line width=1.5]  (26,50) -- (65.35,76.87) ;
\draw [shift={(67,78)}, rotate = 214.32999999999998] [color={rgb, 255:red, 0; green, 0; blue, 0 }  ][line width=0.75]    (10.93,-3.29) .. controls (6.95,-1.4) and (3.31,-0.3) .. (0,0) .. controls (3.31,0.3) and (6.95,1.4) .. (10.93,3.29)   ;
\draw [line width=0.5]  (78,80) -- (142,80) ;
\draw  [shift={(144,80)}, rotate = 180] [color={rgb, 255:red, 0; green, 0; blue, 0 }  ][line width=0.5]    (10.93,-3.29) .. controls (6.95,-1.4) and (3.31,-0.3) .. (0,0) .. controls (3.31,0.3) and (6.95,1.4) .. (10.93,3.29)   ;
\draw  [color={rgb, 255:red, 144; green, 19; blue, 254 }  ,draw opacity=0.6 ][fill={rgb, 255:red, 144; green, 19; blue, 254 }  ,fill opacity=0.15 ] (146,62) -- (152.5,62) -- (152.5,96.43) -- (146,96.43) -- cycle ;
\draw [color={rgb, 255:red, 144; green, 19; blue, 254 }  ,draw opacity=0.6 ][fill={rgb, 255:red, 144; green, 19; blue, 254 }  ,fill opacity=0.15 ] (69,63) -- (75.5,63) -- (75.5,97.43) -- (69,97.43) -- cycle ;
\draw  [dash pattern={on 4.5pt off 4.5pt}]  (27,39) -- (192,39) ;
\draw [shift={(194,39)}, rotate = 180] [color={rgb, 255:red, 0; green, 0; blue, 0 }  ][line width=0.75]    (10.93,-3.29) .. controls (6.95,-1.4) and (3.31,-0.3) .. (0,0) .. controls (3.31,0.3) and (6.95,1.4) .. (10.93,3.29)   ;
\draw  [color={rgb, 255:red, 74; green, 144; blue, 226 }  ,draw opacity=1 ][fill={rgb, 255:red, 74; green, 144; blue, 226 }  ,fill opacity=0.14] (194.38,19.5) -- (201.62,19.5) -- (201.62,76.5) -- (194.38,76.5) -- cycle ;
\draw  [color={rgb, 255:red, 126; green, 211; blue, 33 }  ,draw opacity=1 ][fill={rgb, 255:red, 231; green, 245; blue, 235 }  ,fill opacity=1 ] (235,17.5) -- (242,17.5) -- (242,75.25) -- (235,75.25) -- cycle ;
\draw    [line width=1.5] (203,46) -- (234.5,46.35) ;
\draw [shift={(234.5,46.38)}, rotate = 180.61] [color={rgb, 255:red, 0; green, 0; blue, 0 }  ][line width=1]    (10.93,-3.29) .. controls (6.95,-1.4) and (3.31,-0.3) .. (0,0) .. controls (3.31,0.3) and (6.95,1.4) .. (10.93,3.29)   ;
\draw  [line width=1.5]  (155,79) -- (190.45,50.26) ;
\draw [shift={(192,49)}, rotate = 500.96] [color={rgb, 255:red, 0; green, 0; blue, 0 }  ][line width=0.75]    (10.93,-3.29) .. controls (6.95,-1.4) and (3.31,-0.3) .. (0,0) .. controls (3.31,0.3) and (6.95,1.4) .. (10.93,3.29)   ;
\draw   (60,97) .. controls (60,101.67) and (62.33,104) .. (67,104) -- (100,104) .. controls (106.67,104) and (110,106.33) .. (110,111) .. controls (110,106.33) and (113.33,104) .. (120,104)(117,104) -- (153,104) .. controls (157.67,104) and (160,101.67) .. (160,97) ;
\draw [color={rgb, 255:red, 245; green, 166; blue, 35 }  ,draw opacity=1 ][fill={rgb, 255:red, 255; green, 250; blue, 227 }  ,fill opacity=1 ]
 (298,18.5) -- (305,18.5) -- (305,76.25) -- (298,76.25) -- cycle ;
\draw  [dash pattern={on 0.75pt off 0.75pt}]  (246,47.04) -- (254,47.19) .. controls (255.7,45.56) and (257.37,45.59) .. (259,47.29) .. controls (260.64,48.98) and (262.31,49.01) .. (264,47.38) .. controls (265.7,45.75) and (267.37,45.78) .. (269,47.48) .. controls (270.63,49.18) and (272.29,49.21) .. (273.99,47.58) .. controls (275.68,45.95) and (277.35,45.98) .. (278.99,47.67) .. controls (280.62,49.37) and (282.29,49.4) .. (283.99,47.77) -- (286,47.81) -- (294,47.96) ;
\draw [shift={(296,48)}, rotate = 181.1] [color={rgb, 255:red, 0; green, 0; blue, 0 }  ][line width=0.75]    (10.93,-3.29) .. controls (6.95,-1.4) and (3.31,-0.3) .. (0,0) .. controls (3.31,0.3) and (6.95,1.4) .. (10.93,3.29)   ;
\draw [shift={(244,47)}, rotate = 1.1] [color={rgb, 255:red, 0; green, 0; blue, 0 }  ][line width=0.75]    (10.93,-3.29) .. controls (6.95,-1.4) and (3.31,-0.3) .. (0,0) .. controls (3.31,0.3) and (6.95,1.4) .. (10.93,3.29)   ;

\draw (101,24) node [anchor=north west][inner sep=0.75pt]  [font=\scriptsize] [align=center] {\textit{add}};
\draw (155.13,83.69) node [anchor=north west][inner sep=0.75pt]  [font=\scriptsize,rotate=-318.9] [align=center] {\textit{+ dropout}};
\draw (214.62,80.5) node [anchor=north west][inner sep=0.75pt]  [font=\scriptsize] [align=center] {predictions\\($\displaystyle \mathcal{M}_{i}$)};
\draw (89,112) node [anchor=north west][inner sep=0.75pt]   [align=center] {{\scriptsize hidden layer}};
\draw (7,79.98) node [anchor=north west][inner sep=0.75pt]  [font=\scriptsize] [align=center] {intermed.\\features\\($\displaystyle \mathcal{M}_{i})$};
\draw (277.62,80.98) node [anchor=north west][inner sep=0.75pt]  [font=\scriptsize] [align=center] {predictions\\($\displaystyle \eoe_{i}$)};
\draw (259,30.98) node [anchor=north west][inner sep=0.75pt]  [font=\small] [align=center] {\textbf{$\mathcal{L}_\textit{kd}$}};

\draw  [color={rgb, 255:red, 0; green, 0; blue, 0 }  ,draw opacity=0.15 ][dash pattern={on 5.63pt off 4.5pt}][line width=1]  (1,27.19) .. controls (1,12.72) and (12.73,0.98) .. (27.2,0.98) -- (187.8,0.98) .. controls (203.27,0.98) and (214,12.72) .. (214,27.19) -- (214,105.8) .. controls (214,120.27) and (203.27,132) .. (187.8,132) -- (27.2,132) .. controls (12.73,132) and (1,120.27) .. (1,105.8) -- cycle ;

\draw (60,2.98) node [anchor=north west][inner sep=0.75pt]  [font=\small] [align=center] {\textbf{Adjustment Module}};

\draw (100,67.98) node [anchor=north west][inner sep=0.75pt]  [font=\footnotesize] [align=center] {SiLU};

\end{tikzpicture}

\vspace{-3.8em}
\end{center}
    \vspace{0.2cm}
    \caption{Scheme of the distillation network. Thick solid arrows indicate linear layers. The dashed arrow indicates a skip connection. The network is trained to optimize the knowledge distillation loss between $\mathcal{M}_i$'s predictions and $\eoe_i$'s predictions. }
    \label{fig:distill_diag}
    \vspace{-0.3cm}
\end{figure}
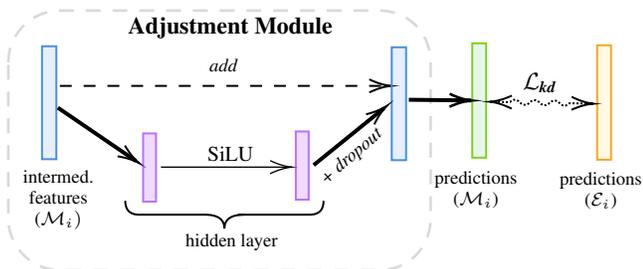

\mysubsubsection{Adjustment module.} One may notice that, if we are to freeze the encoder weights, then the intermediate representations (and, by extension, the input to the ensemble models) are fixed during training. This is a limitation coming from our decision of freezing the encoder weights. We circumvent this by employing a trainable ``adjustment module'' inside each of the experts. This module aims to slightly modify the initial weights, before feeding them into the student classification heads, as well as the ensemble classification heads. 
We choose a simple two-layer neural network architecture as the adjustment model, with a considerably lower number of neurons in the hidden layer than the initial embedding size. The output of the network is then added to the initial embeddings, with dropout applied to the adjustments. 
The motivation for such an architecture is to emulate an adjustment of the embedding vectors $e \in \mathbb{R}^{d \times 1}$ by altering them with a transformation of the form:
\begin{equation}
  e \gets (M + I_d) \cdot e,  
\end{equation}
where $M$ is a low-rank $d \times d$ matrix ($M=UV$, where $U \in \mathbb{R}^{d \times m}$, $V \in \mathbb{R}^{m \times d}$, $m \ll d$). This is only partially accurate, as our model also includes a non-linear activation function and dropout, which makes the adjustment transformation more complex than a simple matrix-vector product. The exact design of the distillation net, with the integrated adjustment module, is shown in
Figure \ref{fig:distill_diag}.

\section{Experimental Evaluation}

We conduct experiments on four action recognition datasets, selecting one expert model per dataset. We carry out the experiments on a computer with an NVIDIA GeForce RTX 3090 GPU, an Intel i9-10940X  3.3 GHz CPU, and 128 GB of RAM.

\subsection{Datasets}

The videos in the four action recognition datasets vary by size, video length, content, and the annotated actions have variable specificity. Some datasets contain similar or even common classes, whereas others are completely disjoint.

\textbf{ActivityNet} \cite{ActivityNet} is a dataset containing $19,994$ untrimmed videos annotated with $200$ activities. The activities 
cover a wide range of complex human activities that are of interest to people in their daily
living. The activity classes are grouped into 7 different high-level categories: Personal Care, Eating and Drinking, Household, Caring and Helping, Working, Socializing
and Leisure, and Sports and Exercises. Unlike other datasets, the videos here have a considerably greater average length of around $2.5$ minutes. The annotations for the test videos are not publicly available. Therefore, following the common practice, we use the validation set along with its publicly-available labels for the final evaluation. Out of the $19,994$ videos, a number of $3,189$ videos were no longer available for download. Therefore, the total size of our dataset is $16,805$. 

\textbf{HMDB51} \cite{HMDB51} is a dataset that consists of $7,000$ clips distributed in $51$ action classes. The classes are more generic compared to the ActivityNet dataset and are grouped in 5 different categories: general facial actions, facial actions with object manipulation, general body movements, body movements with object interaction, body movements for human interaction. The original evaluation scheme uses three different training/testing splits. We consider the first split for our training and evaluation, as the publicly available expert models are trained using the same split. 


\textbf{Kinetics400} \cite{K400} is a large-scale dataset consisting of YouTube videos. It has $400$ action classes, with at least $400$ videos per action. These classes include human-human interactions (e.g.~hugging, shaking hands), as well as human-object interactions (e.g.~playing instruments). Due to its large size, Kinetics400 is often used for model pretraining.

\textbf{UCF101} \cite{UCF101} is a dataset with $13,320$ YouTube videos from $101$ action classes, which vary  in terms of camera motion, object appearance, pose, scale, viewpoint, amount of clutter and illumination. Being of similar complexity with ActivityNet, the $101$ classes belong to 5 types: Body motion, Human-human interactions, Human-object interactions, Playing musical instruments, and Sports. The evaluation scheme could use three different training/testing splits. We use the first split in our experiments.

\subsection{Initial Expert Models}


We use the PyTorch framework, as well as the OpenMMLab's MMAction2 \cite{MMAction2} toolbox, as it provides multiple model architectures pretrained on several datasets. For each dataset, we opt for a pretrained open-source architecture yielding competitive performance in the reported benchmarks, at the time of writing. 

\textbf{ActivityNet.} We use a Temporal Segment Network \cite{TSN} architecture, with a ResNet50 \cite{ResNet} backbone with $8$ segments. As described in the toolbox documentation, the model was pretrained on the Kinetics400 dataset, and then trained for $50$ epochs on the training split of the ActivityNet dataset.

\textbf{HMDB51.} We use a Temporal Shift Module \cite{TSM} architecture, with a ResNet50 \cite{ResNet} backbone with $16$ segments. As described in the MMAction2 toolbox, the model was pretrained on the Kinetics400 dataset, and then trained for $25$ epochs on the first training split of the HMDB51 dataset.

\textbf{Kinetics400.} We use an implementation of the Video Swin Transformer \cite{Swin}, developed on top of the MMAction2 toolbox. More specifically, we use the Swin-B backbone, with a spatial crop of size $244$. As described in the implementation, the model was pretrained on ImageNet-22K.

\textbf{UCF101.} We use the same model as for the HMDB51 case. As described in the MMAction2 toolbox, the model was pretrained on the Kinetics400 dataset, and then trained for $25$ epochs on the first training split of UCF101.





\subsection{Embedding Caching Stage}

\begin{table}[t]
    \centering
    \setlength\tabcolsep{4.5pt}
    \begin{tabular}{|c|c|c|c|}
        \hline
        \textbf{Dataset} & \textbf{Model} & \textbf{Training size} & \textbf{Feature size}  \\
        \hline
        \hline
        ActivityNet & TSN & $8\:398$ & $2048$ \\
        HMDB51 & TSM & $3\:570$ & $2048$ \\
        Kinetics400 & SWIN-B & $226\:070$ & $1024$\\ 
        UCF101 & TSM & $9\:537$ & $2048$ \\
        \hline 
        \multicolumn{2}{|c|}{\textbf{TOTAL}} & $251\:358$ & $7168$ \\
        \hline
    \end{tabular}
    \vspace{0.2cm}
    \caption{Number of training samples per dataset (3rd column), and number of features for each corresponding model (4th column).}
    \label{tab:emb_sizes}
\end{table}

We run inference on the aforementioned models on videos from all datasets to obtain feature vectors from each of the experts. The embedding size obtained for each dataset is shown in Table \ref{tab:emb_sizes}. Ultimately, by concatenating the embeddings, we end up with an aggregated feature vector of size $7168$ for each video from each dataset. We also compute the predicted probabilities for each class alongside the feature vectors.
Both features and predicted probabilities are persisted to disk for quicker experiments. As mentioned earlier, this is a trade-off between the efficiency of the training and the learning potential. 

\subsection{Implementation Details}

\textbf{Teachers.} We model each teacher as a simple ensemble based on a linear meta-classifier. We choose Hinge loss (maximum margin loss) for the ensembles, making each ensemble similar to a soft-margin Support Vector Machines model. We conjecture that an SVM-like architecture would be a robust choice in the context of a high-dimensional system, where the input dimensionality ($7168$) is comparable to and even surpasses the number of examples (see Table \ref{tab:emb_sizes}, \eg HMDB51). 
We also employ a dropout layer before each of the meta-classifiers. We choose separate dropout rates for each model, given by the formula:
\begin{equation}
    p_i = \max\left(0, 1 - \frac{k \cdot \textbf{NumClasses}(\mathcal{D}_i)}{d}\right).
\end{equation}

\begin{table*}[t!]
\setlength\tabcolsep{2.0pt}
    \centering
    \begin{tabular}{|c|ccc|ccc|ccc|ccc|}
    \hline
     \multirow{2}{*}{\textbf{Model}} & \multicolumn{3}{c|}{\textbf{ActivityNet}} & \multicolumn{3}{c|}{\textbf{HMDB51}} & 
     \multicolumn{3}{c|}{\textbf{Kinetics400}} & \multicolumn{3}{c|}{\textbf{UCF101}}  \\
     
     & \texttt{acc@1} & \texttt{acc@5} & \texttt{mAP} 
     & \texttt{acc@1} & \texttt{acc@5} & \texttt{mAP} 
     & \texttt{acc@1} & \texttt{acc@5} & \texttt{mAP}
     & \texttt{acc@1} & \texttt{acc@5} & \texttt{mAP} \\
     
     \hline
     \hline
     
     \textbf{Initial Experts} \cite{TSM, Swin, TSN}
     & $73.81$ & $93.55$ & $42.83$ 
     & $73.60$ & $93.66$ & $61.27$
     & $81.80$ & \textbf{95.19} & \textbf{87.36} 
     & $94.63$ & $99.36$ & $86.32$ \\

     
      \textbf{JEDI Students} \textit{(ours)} 
     & \textbf{81.56} & \textbf{95.31} & \textbf{82.89} 
     & \textbf{75.32} & \textbf{93.92} & \textbf{76.81} 
     & \textbf{82.08} & $94.36$ & $85.94$
     & \textbf{95.97} & \textbf{99.58} & \textbf{97.96} \\ 
     
     \hline
     
     
      \textbf{JEDI Teachers} \textit{(ours)} 
     & 88.50 & $97.42$ & $90.33$ 
     & $78.67$ & $94.90$ & $79.13$ 
     & $81.50$ & $93.98$ & $85.03$ 
     & $98.05$ & $99.84$ & $99.20$ \\ 
     
     
     \hline
     
\end{tabular}
    \vspace{0.2cm}
   \caption{Results of the expert models compared with our  individual students, as well as our teacher ensembles, on ActivityNet, HMDB51, Kinetics400, 
     and UCF101. The most interesting comparison is between the initial experts and our students, since these models are the same. The top scoring individual model is highlighted in bold. Note that the students outperform the initial experts by a large performance gap (e.g. mAP), on all datasets (except Kinetics400). As expected, the teachers (larger ensemble models) outperform the individual students (except Kinetics400).}
    \label{tab:main_results}
\end{table*}

\begin{figure*}[th]
    \begin{center}
        \includegraphics[width=1.0\linewidth]{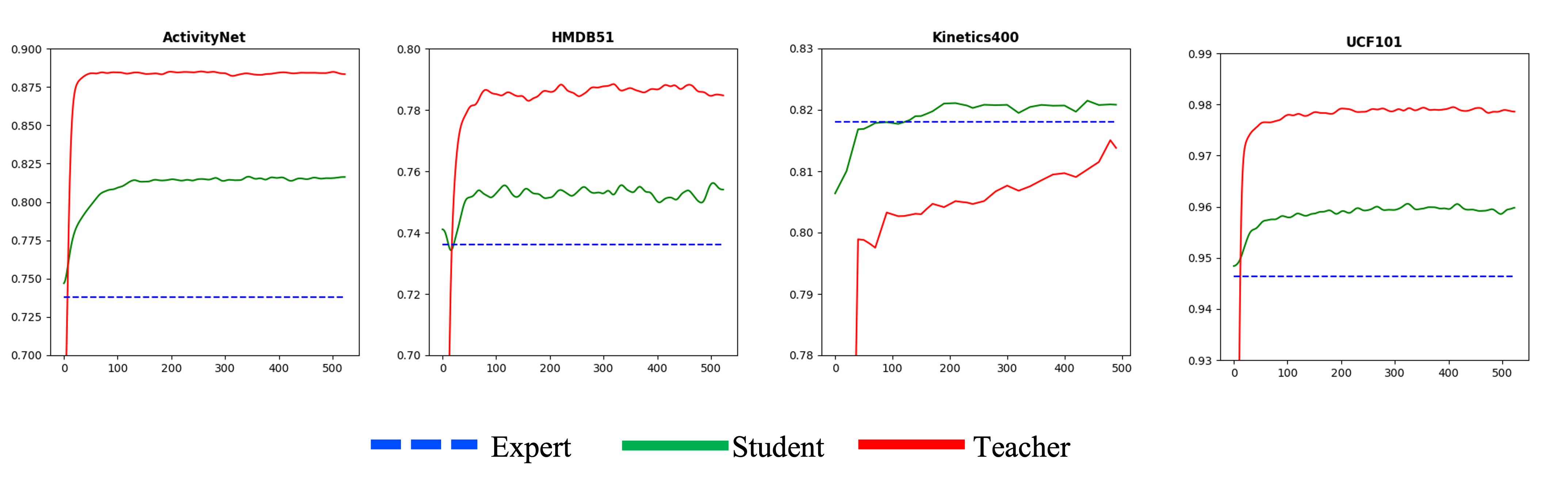} 
        
    \end{center}
    \vspace{-0.4cm}
    \caption{
        \label{fig:loss-graph} 
        Top-1 accuracy evolution over 500 epochs on the test set of each dataset (x-axis shows current epoch, y-axis shows accuracy). JEDI quickly brings both the student and the teacher significantly above the initial experts on all datasets (except for Kinetics400). Best viewed in color.}
        \vspace{-0.2cm}
\end{figure*}

In the above equation, $p_i$ is the dropout probability for ensemble $\mathcal{E}_i$, $d$ is the input dimensionality ($d=7168$). Based on preliminary experiments with $k \in \{1, 10, 100\}$, we find that $k = 10$ is a good choice. Intuitively, the dropout rate is set such that the expected number of active neurons in the training procedure  is $k$ times larger than the number of classes of the given dataset, namely:
\begin{equation}
    \mathbb{E}[\#\text{neurons}]_i = d \cdot (1 - p_i) = k \cdot \textbf{NumClasses}(\mathcal{D}_i).
\end{equation}

In most experiments, we find that the ensembles fit the training data perfectly, obtaining close to $100\%$ accuracy on the training set. However, as the objective of the last SVM layer is to maximize the separation margin between classes, it also proves effective on unseen data.

\textbf{Students.} We keep the encoders of each of the pretrained expert models frozen and we fine-tune only their classification heads. We employ an adjustment module (Figure~\ref{fig:distill_diag}) for each of the experts. The adjustment module consists of a two-layer neural network, with $256$ neurons in the hidden layer (around $12$-$25\%$ of the input dimensionality), and SiLU activation. The adjusted weights are coupled with a Dropout layer with a dropout rate of $0.75$ and combined (via addition) with the original weights.


\textbf{Training setup.}
As mentioned before, in our pipeline, we jointly train the teachers $\eoe_{i}$ and the students $\mathcal{M}_i$ using both supervised classification tasks and unsupervised knowledge distillation tasks. In this sense, we retrain the classification head for all the expert models on the task of predicting the confidence scores of $\eoe_i$.
For knowledge distillation, each student network ($\mathcal{M}_i$) is trained to minimize the cross-entropy loss  between the output predictions and their corresponding teacher's ($\eoe_i$) soft pseudo-label logits with ``\textit{softmax with temperature}'', as popularly suggested by Hinton \etal~\cite{hinton15nipsw}. We set the temperature $T=1$.

We employ the proposed combined loss described in Eq.~\eqref{eq_loss}, choosing $\alpha = \beta = 1$ and $\gamma = 0.4$ for the hyperparameters defined in the equation. 
We employ the AdamW \cite{AdamW} optimizer, with a constant learning rate of $10^{-5}$ and a weight decay factor of $3 \times 10^{-3}$. 
We use a burn-in period of $25$ epochs to warm up the teacher models, before turning on the knowledge distillation losses. Preliminary experiments showed that this warm-up period does not improve the final results, although we have seen slight improvements in convergence speed.
All hyperparameters were chosen after preliminary experiments conducted on a validation set obtained by retaining $15\%$ of the videos from the training set of each dataset $\mathcal{D}_i$. The final experiments and ablation studies are conducted while learning on the entire training sets.
Furthermore, as the hidden features are already pre-computed, the only extra computational cost corresponds to the inference of the non-frozen part of the architecture (the classification heads of each $\mathcal{M}_i$ and the teacher networks $\mathcal{E}_i$). This leads to a very fast training regime. 

\subsection{Results}

We evaluate the performance of our approach in terms of three metrics: top-1 accuracy (\texttt{acc@1}), top-5 accuracy (\texttt{acc@5}), and mean Average Precision (\texttt{mAP}). 
We present quantitative results of our experiments in Table  \ref{tab:main_results}.

%

\mysubsubsection{Performance of students.}
In terms of the top-1 accuracy, {\emph{all}} students learned by distilling the teacher ensembles perform {\emph{better}} than the initial experts, with little extra inference cost (only two extra matrix-vector multiplications). The gains brought by our students are marginal for Kinetics400 and range between $1.34\%$ on UCF101 and $7.75\%$ on ActivityNet. 
In terms of the mAP, we observe the same behavior, with large gains on the three datasets (ActivityNet, HMDB, UCF101), ranging between $11.64\%$ on UCF101 and $40.06\%$ on ActivityNet. 


\begin{table*}[t]
    \centering
    \setlength\tabcolsep{3.5pt}
    \begin{tabular}{|c|c c|c c|c c|}
    \hline
    \multirow{2}{*}{\textbf{Scenario}} & \multicolumn{2}{c|}{\textbf{ActivityNet}} & \multicolumn{2}{c|}{\textbf{HMDB51}} & \multicolumn{2}{c|}{\textbf{UCF101}}  \\
    & \textit{Teacher} & \textit{Student} & \textit{Teacher} & \textit{Student} & \textit{Teacher} & \textit{Student} \\
    \hline
    \hline
    
    \textit{(Initial Expert)} & -- & $73.81$ & -- & $73.60$ & -- & $94.63$ \\
    
    \hline
    
    \textbf{Predictions} & $85.59$ & $80.40$ & $73.59$ & $72.10$ & $97.30$ & $94.86$  \\
    
    \textbf{Base Features} 
    & $88.27$ & \textbf{81.65} 
    & $78.13$ & $74.25$ 
    & $98.02$ & \textbf{96.04} \\

    \textbf{Adjusted Features} 
    & \textbf{88.50} & $81.56$ 
    & {78.67} & \textbf{75.32}   
    & \textbf{98.05} & $95.97$  \\

    \textbf{Adjusted Features + Predictions} 
        & $87.93$ & $81.15$ 
        & \textbf{78.71} & $74.80$ 
        & $97.76$ & $95.62$ \\
    
    \hline
    
    \textbf{No Distillation} &  $87.79$ & $71.74$ & $77.55$ & $71.95$ & $97.71$ & $95.52$ \\
    
    \textbf{Single Dataset} 
    & $88.17$ & $78.86$ 
    & $78.09$ & $73.43$ 
    & $97.71$ & $95.44$ \\
    
    \textbf{Just Kinetics} 
    & \textbf{88.57} & $80.81$ 
    & $77.84$ & $73.98$   
    & $97.76$ & $95.81$  \\
     
    \textbf{All Datasets} 
    & {88.50} & \textbf{81.56} 
    & \textbf{78.67} & \textbf{75.32}   
    & \textbf{98.05} & \textbf{95.97}  \\
    
    \hline
     
\end{tabular}
    

    \vspace{0.2cm}
         \caption{Top-1 accuracy rates of ablation study scenarios. First part indicates the starting expert models. Second part describes scenarios with varying ensemble input strategies (as described in \ref{sec:ens_input}). Third part describes scenarios with varying unsupervised distillation strategies (as described in \ref{sec:distill}). Results are reported on ActivityNet, HMDB51, UCF101. The best results in each section are highlighted in bold.}
    \label{tab:abl_results}
\end{table*}

\mysubsubsection{Performance of teachers.} For completeness, we also show the performance of the final teachers on each dataset. 

Unsurprisingly, our ensembles of experts perform{\emph{ much better}} than the initial experts on all chosen datasets apart from Kinetics400. The trained teachers manage to surpass the initial experts with accuracy margins as high as $14\%$ (on the ActivityNet dataset), and the improvements are consistent across datasets. 


We observe that the experiments on the Kinetics400 dataset seem to be a special case, probably due to the very large size of the dataset when compared to others (as shown in  Table~\ref{tab:emb_sizes}). As before, we observe an improvement over semi-supervised iterations for both teacher and student, but the teacher remains weaker than the initial expert, probably due to the additional features from the other small datasets which, in this case, seem to reduce the generalization power. However, what is still a positive result is that the additional pseudo-labels provided by the teacher continue to help the student model, which outperforms the initial expert. Moreover, the fact that the student outperforms the more complex teacher is in fact a pleasant surprise, but this less common case is not unheard of in the literature. It is in fact known that when teachers tend to overfit (it seems to be our case on Kinetics400), the simpler students can generalized better, outperforming their teachers~\cite{croitoru2021teachtext}.




\mysubsubsection{Improvements during training.}
Figure \ref{fig:loss-graph} shows the improvement of the teachers and students, over training iterations (epochs), in terms of top-1 accuracy. The training process displays fast convergence for all datasets, except for the much larger Kinetics400. 
The students initially obtain better performance than the teacher ensembles (due to their pretraining), but the teachers quickly catch up. The sudden performance drop at epoch $25$ is due to the introduction of knowledge distillation into the pipeline (which proves effective shortly after). 
The plots show the accuracy on test data, which could explain the fluctuation for the teacher on the Kinetics400 dataset. However, we notice that the teacher improves and it could potentially overcome the initial expert if left to train for more epochs. By design, we chose to stop the training at 500 epochs for all datasets, and as it can be seen the experiments prove in all cases the benefit of our semi-supervised approach. Each student improves significantly over the corresponding initial expert, while the teachers exhibit considerable accuracy gains in three out of four cases.

\section{Ablation Study}

In order to fully motivate our claims and better understand the source of the significant improvements in our methods, we propose several ablation experiments.
We restrict ourselves to three of the four datasets and exclude Kinetics400, as the experiments are much more time consuming to run on this dataset. 

\subsection{\label{sec:ens_input}Input to Ensembles}

We test the impact of different input features for our ensemble of experts forming our pipeline. We analyze four different scenarios: 
\begin{enumerate*}
    \item \textbf{Predictions:} We use the \textit{prediction logits} of the refined experts as input to the ensembles.
    \item \textbf{Base Features:} We use the intermediate features \textit{before the adjustment module} as input to the ensembles. As these features are fixed during training, in this scenario, the teachers are completely decoupled from the students (the dependence is \textit{unidirectional}).
    \item \textbf{Adjusted Features}. We use the intermediate features \textit{after the adjustment module} as input to the ensembles. As these features change during training, the teachers end up being jointly trained with the students (the dependence is \textit{bidirectional}). This is the proposed scenario presented in the main results, corresponding to our final model.
    \item \textbf{Adjusted Features + Predictions}. We use the adjusted features, along with predictions as input to the ensembles.
\end{enumerate*}

We compute the top-1 accuracy of the teachers and the distilled students on the testing set of their corresponding dataset. The results are reported in the second section of Table \ref{tab:abl_results}. The results show that feeding just predictions as input yields no overall improvement over the experts on two out of three datasets. We also observe a slight difference in performance between using base features and adjusted features, in terms of teacher performance. Interestingly, we find that the continuous improvement of the students mostly affects the future generation teachers' performance, and it hardly bootstraps back to the children themselves. Surprisingly, we see no improvement (even some form of degradation) when combining predictions with hidden features as opposed to using just hidden features. This goes against the naive intuition that more input features makes for a better network.

By contrasting the performance reports in the \textbf{Predictions}, \textbf{Adjusted Features}, and \textbf{Adjusted Features+Predictions} scenarios, we conclude that: \textbf{1)} The output predictions of experts alone leverage considerably less knowledge than using the intermediate features; and \textbf{2)} Predictions do not add extra information when intermediate features are present.
In fact, one may see that predictions are ultimately a simple (linear, even) function of the intermediate features, \ie: $$\textit{predictions}=\textbf{ClassificationHead}(\textit{features}).$$

\subsection{\label{sec:distill} Effectiveness of Unsupervised Distillation}

We claim that the choice of using data from all datasets in a semi-supervised scenario is beneficial to the effective distillation of the teachers back into the students. We propose an ablation study to empirically test this hypothesis. Our study analyzes three scenarios: 

\begin{enumerate*}
    \item \textbf{No Distillation}. Both students and teachers are trained only on the supervised classification tasks. 
    \item \textbf{Single Dataset}. Each student uses only data from the training set of its original dataset for distillation. 
     \item \textbf{Just Kinetics}. We take a sample of $20,000$ videos from the Kinetics400 dataset for distillation. This data is only used for unsupervised training (the labels are merely ignored).
    \item \textbf{All Datasets}. The students are fed with data from the training sets of all datasets for distillation. This is the scenario corresponding to the main results. 
    
\end{enumerate*}

Table \ref{tab:abl_results} shows the top-1 accuracy for students and teachers.
First, if no distillation is performed at all (the \textbf{No Distillation} scenario), the single expert performance degrades for two out of three models. We explain this phenomenon by stating that, in this scenario, the single experts are trained with supervision on the data already available for them during pretraining, and no common knowledge is leveraged. Moreover, in our training procedure, we do not employ techniques that were used during pretraining (in particular, data augmentation techniques), which ultimately lead to experts overfitting the (now unaugmented) data. 

In the \textbf{Single Dataset} case, the performance of students is improved on two out of three datasets. It is also important that the performance of the teacher ensembles improves as well. This shows that the unsupervised training stage is effective and the improvement of students positively influences the combined ensemble teachers. Using \textbf{Just Kinetics} data, however, suffers from the out-of-distribution bias of the Kinetics400 dataset. This is especially visible on HMDB51, where the labels are much more semantically different from Kinetics400 than from the other datasets. The \textbf{All Data} scenario (main results) benefits from both the inside-distribution data of the own dataset and the out-of-distribution data of the other datasets, and its performance is notably higher than the other scenarios. This indicates that the extra data provided from the other datasets helps improve the robustness and generalization power of each student net. Overall, the ablation study shows that the inclusion of knowledge distillation combined with our joint approach is highly effective.

\section{Conclusions and Future work}

In this work, we proposed JEDI, a semi-supervised learning model that distills knowledge learned across several datasets into student models corresponding to each dataset. The process takes the form of a student-teacher paradigm. Teachers are ensembles of previous generation students, which provide supervisory signals to the next generation students, in a self-supervised distillation fashion. Learning of students and teachers is performed jointly, by employing a novel cost function. Our extensive experiments demonstrated that our approach efficiently improves the students, as well as the teachers (even after a few learning epochs), significantly outperforming the original experts by a wide margin on challenging datasets. The practical advantage at test time is significant, since the highly improved students have the same inference cost as the initial experts. 

Although we have focused on the problem of action classification in video, by covering vastly different types of actions and contexts, the overall approach is quite general and could be used in many other multi-dataset multi-task scenarios. In future work, we plan to further explore and develop our approach in the realm of other domains.

\subsection*{Acknowledgment}
This work was funded in part by UEFISCDI, under Projects EEA-RO-2018-0496 and PN-III-P4-ID-PCE-2020-2819.

{\small
\bibliographystyle{ieee_fullname}
\bibliography{egbib}

\begin{thebibliography}{10}\itemsep=-1pt

\bibitem{ba14nips}
Jimmy Ba and Rich Caruana.
\newblock Do deep nets really need to be deep?
\newblock In {\em Proceedings of NIPS}, volume~27. Curran Associates, Inc.,
  2014.

\bibitem{caron2020unsupervised}
Mathilde Caron, Ishan Misra, Julien Mairal, Priya Goyal, Piotr Bojanowski, and
  Armand Joulin.
\newblock Unsupervised learning of visual features by contrasting cluster
  assignments.
\newblock In {\em Proceedings of NeurIPS}, volume~33, pages 9912--9924, 2020.

\bibitem{Carreira_2017_CVPR}
Joao Carreira and Andrew Zisserman.
\newblock {Quo Vadis, Action Recognition? A New Model and the Kinetics
  Dataset}.
\newblock In {\em Proceedings of CVPR}, July 2017.

\bibitem{MMAction2}
MMAction2 Contributors.
\newblock Openmmlab's next generation video understanding toolbox and
  benchmark.
\newblock {https://github.com/open-mmlab/mmaction2}, 2020.

\bibitem{croitoru2021teachtext}
Ioana Croitoru, Simion-Vlad Bogolin, Marius Leordeanu, Hailin Jin, Andrew
  Zisserman, Samuel Albanie, and Yang Liu.
\newblock {TEACHTEXT: CrossModal Generalized Distillation for Text-Video
  Retrieval}.
\newblock In {\em Proceedings of ICCV}. IEEE, 2021.

\bibitem{deng2009imagenet}
Jia Deng, Wei Dong, Richard Socher, Li-Jia Li, Kai Li, and Li Fei-Fei.
\newblock {ImageNet: A large-scale hierarchical image database}.
\newblock In {\em Proceedings of CVPR}, pages 248--255, 2009.

\bibitem{doersch2017multitask}
Carl Doersch and Andrew Zisserman.
\newblock Multi-task self-supervised visual learning.
\newblock In {\em Proceedings of ICCV}, 2017.

\bibitem{bigbigan}
Jeff Donahue and Karen Simonyan.
\newblock Large scale adversarial representation learning.
\newblock In {\em {Proceedings of NeurIPS}}, 2019.

\bibitem{Douze11cvpr}
Matthijs Douze, Arnau Ramisa, and Cordelia Schmid.
\newblock Combining attributes and fisher vectors for efficient image
  retrieval.
\newblock In {\em {Proceedings of CVPR}}, pages 745--752. {IEEE} Computer
  Society, 2011.

\bibitem{Feichtenhofer2016CVPR}
Christoph Feichtenhofer, Axel Pinz, and Andrew Zisserman.
\newblock Convolutional two-stream network fusion for video action recognition.
\newblock In {\em Proceedings of CVPR}, June 2016.

\bibitem{Georgescu-CVPR-2021}
Mariana-Iuliana Georgescu, Antonio Barbalau, Radu~Tudor Ionescu, Fahad~Shahbaz
  Khan, Marius Popescu, and Mubarak Shah.
\newblock {Anomaly Detection in Video via Self-Supervised and Multi-Task
  Learning}.
\newblock In {\em Proceedings of CVPR}, pages 12742--12752, 2021.

\bibitem{predict-rotation}
Spyros Gidaris, Praveer Singh, and Nikos Komodakis.
\newblock Unsupervised representation learning by predicting image rotations.
\newblock In {\em {Proceedings of ICLR}}, 2018.

\bibitem{Gowda2021AAAI}
Shreyank~N Gowda, Marcus Rohrbach, and Laura Sevilla-Lara.
\newblock Smart frame selection for action recognition.
\newblock In {\em Proceedings of AAAI}, volume~35, pages 1451--1459, May 2021.

\bibitem{Hansen1990pami}
L.~K. Hansen and P. Salamon.
\newblock Neural network ensembles.
\newblock {\em IEEE Transactions on Pattern Analysis and Machine Intelligence},
  pages 993--1001, 1990.

\bibitem{Hara_2018_CVPR}
Kensho Hara, Hirokatsu Kataoka, and Yutaka Satoh.
\newblock {Can Spatiotemporal 3D CNNs Retrace the History of 2D CNNs and
  ImageNet?}
\newblock In {\em Proceedings of CVPR}, June 2018.

\bibitem{ResNet}
Kaiming He, Xiangyu Zhang, Shaoqing Ren, and Jian Sun.
\newblock Deep residual learning for image recognition.
\newblock In {\em Proceedings of CVPR}, pages 770--778, 2016.

\bibitem{ActivityNet}
Fabian~Caba Heilbron, Victor Escorcia, Bernard Ghanem, and Juan~Carlos Niebles.
\newblock Activitynet: A large-scale video benchmark for human activity
  understanding.
\newblock In {\em Proceedings of CVPR}, pages 961--970, 2015.

\bibitem{hinton15nipsw}
Geoffrey Hinton, Oriol Vinyals, and Jeffrey Dean.
\newblock Distilling the knowledge in a neural network.
\newblock In {\em NIPS Deep Learning and Representation Learning Workshop},
  2015.

\bibitem{K400}
Will Kay, Joao Carreira, Karen Simonyan, Brian Zhang, Chloe Hillier, Sudheendra
  Vijayanarasimhan, Fabio Viola, Tim Green, Trevor Back, Paul Natsev, et~al.
\newblock {The Kinetics Human Action Video Dataset}.
\newblock {\em arXiv preprint arXiv:1705.06950}, 2017.

\bibitem{kemker2018measuring}
Ronald Kemker, Marc McClure, Angelina Abitino, Tyler Hayes, and Christopher
  Kanan.
\newblock Measuring catastrophic forgetting in neural networks.
\newblock In {\em Proceedings of AAAI}, volume~32, 2018.

\bibitem{HMDB51}
H. Kuehne, H. Jhuang, E. Garrote, T. Poggio, and T. Serre.
\newblock {HMDB}: a large video database for human motion recognition.
\newblock In {\em Proceedings of ICCV}, 2011.

\bibitem{Leordeanu21aaai}
Marius Leordeanu, Mihai~Cristian P{\^{\i}}rvu, Dragos Costea, Alina~Elena
  Marcu, Emil Slusanschi, and Rahul Sukthankar.
\newblock Semi-supervised learning for multi-task scene understanding by neural
  graph consensus.
\newblock In {\em {Proceedings of AAAI}}, pages 1882--1892. {AAAI} Press, 2021.

\bibitem{li2018resound}
Yingwei Li, Yi Li, and Nuno Vasconcelos.
\newblock Resound: Towards action recognition without representation bias.
\newblock In {\em Proceedings of ECCV}, pages 513--528, 2018.

\bibitem{Li_2019_CVPR}
Yi Li and Nuno Vasconcelos.
\newblock {REPAIR: Removing Representation Bias by Dataset Resampling}.
\newblock In {\em Proceedings of CVPR}, June 2019.

\bibitem{TSM}
Ji Lin, Chuang Gan, and Song Han.
\newblock {TSM: Temporal Shift Module for Efficient Video Understanding}.
\newblock In {\em Proceedings of ICCV}, 2019.

\bibitem{lin2014microsoft}
Tsung-Yi Lin, Michael Maire, Serge Belongie, James Hays, Pietro Perona, Deva
  Ramanan, Piotr Dollar, and Larry Zitnick.
\newblock {Microsoft COCO: Common Objects in Context}.
\newblock In {\em Proceedings of ECCV}, September 2014.

\bibitem{Liu_2021_CVPR}
Xin Liu, Silvia~L. Pintea, Fatemeh~Karimi Nejadasl, Olaf Booij, and Jan~C. van
  Gemert.
\newblock No frame left behind: Full video action recognition.
\newblock In {\em Proceedings of CVPR}, pages 14892--14901, June 2021.

\bibitem{Liu2019a}
Y. Liu, S. Albanie, A. Nagrani, and A. Zisserman.
\newblock Use what you have: Video retrieval using representations from
  collaborative experts.
\newblock In {\em Proceedings of BMVC}, 2019.

\bibitem{Swin}
Ze Liu, Yutong Lin, Yue Cao, Han Hu, Yixuan Wei, Zheng Zhang, Stephen Lin, and
  Baining Guo.
\newblock Swin transformer: Hierarchical vision transformer using shifted
  windows.
\newblock In {\em Proceedings of ICCV}, 2021.

\bibitem{AdamW}
Ilya Loshchilov and Frank Hutter.
\newblock Decoupled weight decay regularization.
\newblock {\em arXiv preprint arXiv:1711.05101}, 2017.

\bibitem{lu2020corr}
Yao Lu, Sören Pirk, Jan Dlabal, Anthony Brohan, Ankita Pasad, Zhao Chen,
  Vincent Casser, Anelia Angelova, and Ariel Gordon.
\newblock {Taskology: Utilizing Task Relations at Scale}.
\newblock In {\em Proceedings of CVPR}, pages 8696--8705, 2021.

\bibitem{shuffle-herbert}
Ishan Misra, Lawrence~C. Zitnick, and Martial Hebert.
\newblock Shuffle and learn: Unsupervised learning using temporal order
  verification.
\newblock In {\em {Proceedings of ECCV}}, 2016.

\bibitem{Ren_2020_WACV}
Jian Ren, Xiaohui Shen, Zhe Lin, and Radomir Mech.
\newblock Best frame selection in a short video.
\newblock In {\em Proceedings of WACV}, March 2020.

\bibitem{Simonyan2014nips}
Karen Simonyan and Andrew Zisserman.
\newblock Two-stream convolutional networks for action recognition in videos.
\newblock In {\em Proceedings of NIPS}, volume~27. Curran Associates, Inc.,
  2014.

\bibitem{Singh_2016_CVPR}
Bharat Singh, Tim~K. Marks, Michael Jones, Oncel Tuzel, and Ming Shao.
\newblock A multi-stream bi-directional recurrent neural network for
  fine-grained action detection.
\newblock In {\em Proceedings of CVPR}, June 2016.

\bibitem{UCF101}
Khurram Soomro, Amir~Roshan Zamir, and Mubarak Shah.
\newblock {UCF101: A dataset of 101 human actions classes from videos in the
  wild}.
\newblock {\em arXiv preprint arXiv:1212.0402}, 2012.

\bibitem{cluster-herbert}
Pavel Tokmakov, Martial Hebert, and Cordelia Schmid.
\newblock Unsupervised learning of video representations via dense trajectory
  clustering.
\newblock In {\em Proceedings of ECCVW}, pages 404--421. Springer, 2020.

\bibitem{TommasiPCT17}
Tatiana Tommasi, Novi Patricia, Barbara Caputo, and Tinne Tuytelaars.
\newblock A deeper look at dataset bias.
\newblock In {\em Domain Adaptation in Computer Vision Applications}, Advances
  in Computer Vision and Pattern Recognition, pages 37--55. Springer, 2017.

\bibitem{Torralba11unbiasedlook}
Antonio Torralba and Alexei~A. Efros.
\newblock Unbiased look at dataset bias.
\newblock In {\em Proceedings of CVPR}, 2011.

\bibitem{TSN}
Limin Wang, Yuanjun Xiong, Zhe Wang, Yu Qiao, Dahua Lin, Xiaoou Tang, and Luc
  Van~Gool.
\newblock Temporal segment networks: Towards good practices for deep action
  recognition.
\newblock In {\em Proceedings of ECCV}, pages 20--36. Springer, 2016.

\bibitem{zamir2016generic}
Amir~R Zamir, Tilman Wekel, Pulkit Agrawal, Colin Wei, Jitendra Malik, and
  Silvio Savarese.
\newblock Generic {3D} representation via pose estimation and matching.
\newblock In {\em Proceedings of ECCV}, pages 535--553. Springer, 2016.

\bibitem{unsup-clust-cvpr}
Xiaohang Zhan, Jiahao Xie, Ziwei Liu, Yew{-}Soon Ong, and Chen~Change Loy.
\newblock Online deep clustering for unsupervised representation learning.
\newblock In {\em {Proceedings of CVPR}}, 2020.

\bibitem{colorization}
Richard Zhang, Phillip Isola, and Alexei~A. Efros.
\newblock Colorful image colorization.
\newblock In {\em {Proceedings of ECCV}}, 2016.

\bibitem{split-brain}
Richard Zhang, Phillip Isola, and Alexei~A. Efros.
\newblock {Split-Brain Autoencoders: Unsupervised Learning by Cross-Channel
  Prediction}.
\newblock In {\em {Proceedings of CVPR}}, 2017.

\end{thebibliography}
}

\end{document}